# Temporal Fusion Transformers for Streamflow Prediction: Value of Combining Attention with Recurrence

Sinan Rasiya Koya[1] and Tirthankar Roy[1]

[1]Department of Civil and Environmental Engineering, University of Nebraska-Lincoln, USA

## Abstract

Over the past few decades, the hydrology community has witnessed notable advancements in streamflow prediction, particularly with the introduction of cutting-edge machine-learning algorithms. Recurrent neural networks, especially Long Short-Term Memory (LSTM) networks, have become popular due to their capacity to create precise forecasts and realistically mimic the system dynamics. Attention-based models, such as Transformers, can learn from the entire data sequence concurrently, a feature that LSTM does not have. This work tests the hypothesis that combining recurrence with attention can improve streamflow prediction. We set up the Temporal Fusion Transformer (TFT) architecture, a model that combines both of these aspects and has never been applied in hydrology before. We compare the performance of LSTM, Transformers, and TFT over 2,610 globally distributed catchments from the recently available Caravan dataset. Our results demonstrate that TFT indeed exceeds the performance benchmark set by the LSTM and Transformers for streamflow prediction. Additionally, being an explainable AI method, TFT helps in gaining insights into the streamflow generation processes.

## 1. Introduction

Recently, recurrent neural networks (RNN) have been widely explored in streamflow prediction problems [1–4]. RNNs are particularly useful for learning patterns in sequential (or time-series) data [5,6]. With nodes connected cyclically, RNNs are trained to retain information from previous data points. Long Short-Term Memory (LSTM; Hochreiter and Schmidhuber, 1997) networks, an widely applied RNN in hydrology, can learn from both the short-term and long-term temporal dynamics of hydrological processes in a basin through its internal memory states [1]. One of the first applications of LSTM in hydrology was by Fang et al. (2017), where the authors used LSTM to predict the Soil Moisture Active Passive (SMAP) level-3 moisture product with atmospheric forcings[8]. Later, Kratzert et al. (2018) used LSTM for streamflow prediction, showing that they can outperform the predictions from popular hydrological models[1]. One of the main drawbacks of LSTM (or any RNNs) is that the model processes data step-by-step, where the current states are solely calculated from previous states, with no direct access to distant historical data. Many studies have shown that, regardless of the information retained in memory states, LSTM often fails to learn from long-term patterns [9,10]. Thus, LSTM is most likely not learning significantly from the long-term seasonal patterns in the hydrological system in streamflow modeling. This is where attention-based models can be a potential candidate for streamflow prediction.

Transformers [9], a novel sequence modeling architecture based on the attention mechanism, created a new wave, not just in sequence modeling but in the entire field of machine learning. One of the critical features of Transformer architecture, unlike traditional RNNs, is that it can directly access the entire sequence. Leveraging the multi-head attention mechanism, Transformers can learn from complete sequences, including distant and near-past information [9]. This potentially makes Transformers a better model for studying long-term seasonal patterns in the data. Moreover, Transformers can predict multiple steps of output sequence at a time, which can be tricky in the case of LSTM (or any RNNs). This property is handy for multi-step streamflow prediction. Currently, there are very few applications of Transformer models in hydrology. Yin et al. (2022) proposed a rainfall-runoff modeling framework based on Transformer architecture[11]. They showed that the attention-based model predicts better than the benchmark LSTM for catchments in the US. Castangia et al. (2023) applied the encoder (see section 2.3.3) part of a Transformer for flood forecasting in a single basin[12]. Although a Transformer network has the above-mentioned advantages, it does not represent the dynamic nature rooted in a hydrologic system. The interpretation, such that taking the LSTM cell states analogous to storage layers, cannot be made with Transformers. For that reason, combining recurrence (as in RNNs) and attention (as in Transformers) can potentially improve the streamflow prediction.

Lim et al. (2021) introduced Temporal Fusion Transformers (TFT) which incorporate the recurrent aspects of LSTMs and attention mechanisms of Transformers[13]. TFT architecture is developed by efficiently aligning LSTM and multi-head attention mechanisms with several other components to produce the best prediction. TFTs have shown significant improvement over benchmarks in time-series predictions [13]. TFT has considerable merits over LSTM and Transformers. It integrates static covariates with continuous variables. TFTs can be interpreted (unlike the black-box nature of LSTM and Transformers) with the help of the variable selection networks, which identify relevant input features, and attention scores given by the model for past time steps. To the best of our knowledge, this study is the first one to implement TFTs in the field of hydrology.

We hypothesize that a combination of attention with recurrence could improve streamflow predictions. To test our hypothesis, we trained TFTs, Transformers, and LSTM separately for all Caravan [14] basins that do not have significant missing data. We compared the prediction performance of the three models. Our results show how TFTs can be a better candidate for streamflow prediction problems. The subsequent sections in this article discuss the details of methods used and results, followed by concluding remarks.

## 2. Methods

### 2.1. Study Area and Data

As we aim to implement the three models (TFT, Transformers, and LSTM) on basin aggregated data, we used the Caravan dataset in this study [14]. Caravan is an open-source dataset that provides daily streamflow data, meteorological forcing, and static attributes of 6,830 basins worldwide. This dataset is derived through standardizing Catchment Attributes and MEteorology for Large-sample Studies (CAMELS) datasets existing for multiple regions (US, Great Britain, Brazil, Australia, and Chile), the Hydrometeorological Sandbox - École de technologie supérieure (HYSETS), and LArge-SaMple DAta for Hydrology and Environmental Sciences for Central Europe (LamaH-CE) [15–21]. The meteorological forcing in the dataset includes selected variables that are important for hydrology, aggregated from ERA5-Land data [22]. The static catchment attributes are gathered from HydroATLAS [23] and daily time series derived from ERA5-Land (for more details, see Kratzert et al., 2023). In total, we use 38 continuous and 211 static variables. We removed the basins which have continuously 30 days or more of missing data. The remaining basins with missing data are filled by cubic spline interpolation. From this interpolated set of basins, we removed the ones that had less than 30 years of data, which gave us 2,610 basins across the globe (Figure 1). To the best of our knowledge, our study is the first to implement and compare three state-of-the-art machine learning models, as discussed in the following sections, on such a wide range of basins globally.

### 2.2. LSTM

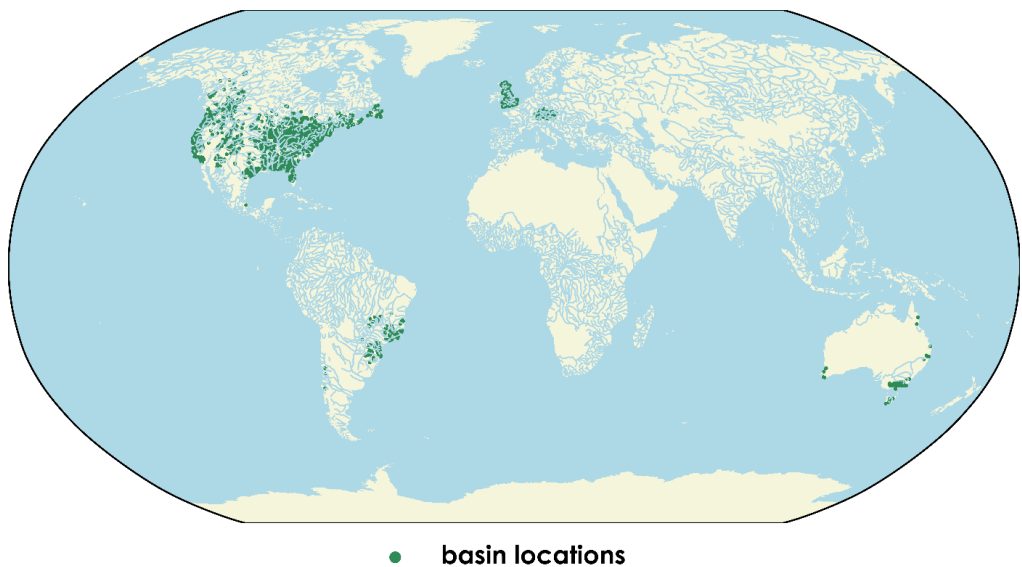

**Figure 1**. Locations of Caravan basins included in this study.

Long Short-Term Memory (LSTM; Figure 2) is a widely used variety of recurrent neural networks applied to streamflow prediction [4,24–26]. The memory cell state inside LSTM, arguably, helps to learn the long-term dependencies in sequential data, which makes it suitable for modeling dynamical systems such as basin hydrology [1]. In a forward pass, an LSTM cell executes the following equations.

$$i_t = \sigma(W_i x_t + U_i h_{t-1} + b_i) \tag{1}$$

$$f_t = \sigma(W_f x_t + U_f h_{t-1} + b_f) \tag{2}$$

$$o_t = \sigma(W_o x_t + U_o h_{t-1} + b_o) \tag{3}$$

$$c_t = f_t \odot c_{t-1} + i_t \odot tanh(W_g x_t + U_g h_{t-1} + b_g) \tag{4}$$

$$h_t = o_t \odot tanh(c_t) \tag{5}$$

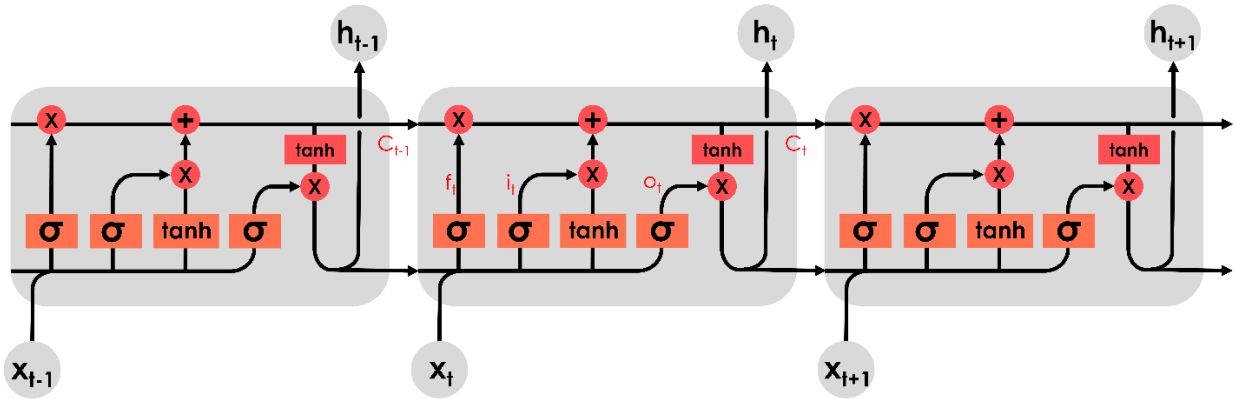

**Figure 2.** Structure of an LSTM cell unrolled for three time steps.

Here, $x_t$ is the input at time step $t$ and $h_t$ is the hidden state at time step $t$. $\odot$ represents the element-wise multiplication, $tanh$ is the hyperbolic tangent activation function, and σ represents the sigmoid activation function. W, U, and b are the weights and biases that are optimized during training. The cell states ($c_t$ and $h_t$) of time step $t$ is used in the processing input of the next time step. $c_t$ is the memory state retaining long-term information and $h_t$ is the memory state retaining short-term information. In this study, we implemented LSTM networks with three layers and with ten nodes in the hidden layers. The output of the final time step is passed through a linear layer and sigmoid activation function followed by another linear layer mapping to the target streamflow.

## 2.3. Transformers

Vaswani et al. (2017) introduced Transformers for sequence modeling by leveraging the attention mechanism [9,27]. Amongst the key features of Transformers is that it can "attend" and learn from the entire sequence at once, one that LSTM cannot. A Transformer architecture (Figure 3a) typically consists of four major components (1) positional encoding, (2) attention mechanism, (3) encoder block, and (4) decoder block.

### 2.3.1. Positional Encoding

A Transformer model uses positional encoding to learn the relative position of data in the input sequence. In a regular RNN, such as LSTM, the model structure contains information about the relative positions inside a sequence. This is evident from the unrolled structure of LSTMs (Figure 2). Unlike RNNs, the Transformers do not have any recurrence in their structure. Therefore, there is a need for a mechanism to inform the model about the relative positioning of data inside a sequence, which is what the positional encoder does. This is accomplished by adding PE from Equations 6 and 7 with the input sequence linearly transformed to a dimension of $d_{model}$ (we use 16).

$$PE_{(pos,2i)} = sin\left(\frac{pos}{10000^{2i/d_{model}}}\right) \tag{6}$$

$$PE_{(pos,2i+1)} = cos\left(\frac{pos}{10000^{2i/d_{model}}}\right) \tag{7}$$

Here, *pos* is the relative position (1 to $d_{model}$ in our case) and *i* is the index of the linearly transformed input sequence (1 to 249, total number of variables) at position *pos*.

### 2.3.2. Attention

The attention mechanism was introduced in machine translation problems to automatically identify relevant information from any part of a sequence [27]. In a Transformer, we use multi-head self-attention. The entire input sequence is passed through a linear layer creating three matrices, which are called queries (Q), keys (K), and values (V), respectively. Single-head attention can be represented with Equation 8, also known as scaled dot-product attention.

$$Attention(Q, K, V) = softmax\left(\frac{QK^T}{\sqrt{d_k}}\right)V \tag{8}$$

In an *n*-head self-attention (we use four heads), Equation 8 is separately applied, concatenated together, and passed through a linear layer producing sequences of the same size as that of decoder input (Figure 3b and 3c).

### 2.3.3. Encoder

An encoder block has two components: a multi-head attention layer that takes queries, keys, and values and a feed-forward network layer. In both of these layers, their inputs are added to the outputs (residual connection), and layer normalization is carried out [28]. We can have multiple encoder blocks stacked one after another. In this study, we used two blocks. Note that we use masked multi-head attention in the encoder block. This is because, unlike the problems in Natural Language Processing, a discharge of today can never be influenced by any future states of input variables. In a masked multi-head attention layer, we add an upper triangular matrix of -Inf to scaled $QK^T$.

### 2.3.4. Decoder

A decoder block is similar to an encoder block, with an additional masked multi-head attention layer before the regular multi-head attention (Figure 3a). The queries and keys to the multi-head attention layer of a decoder come from the encoder output, and the values come from a masked multi-head attention layer. The input to a decoder block is the positional encoded variables of the previous step (here, the previous day). Since we only predict one time-step, we only provide one time-step data into the decoder as it should be of the same length of output sequence [9]. As a consequence, the masking in the attention layer is inactive in our model. Same as encoder blocks, we can stack decoder blocks one after another. In our model, we stacked three decoder blocks followed by a linear layer mapping to seven quantiles of the target streamflow.

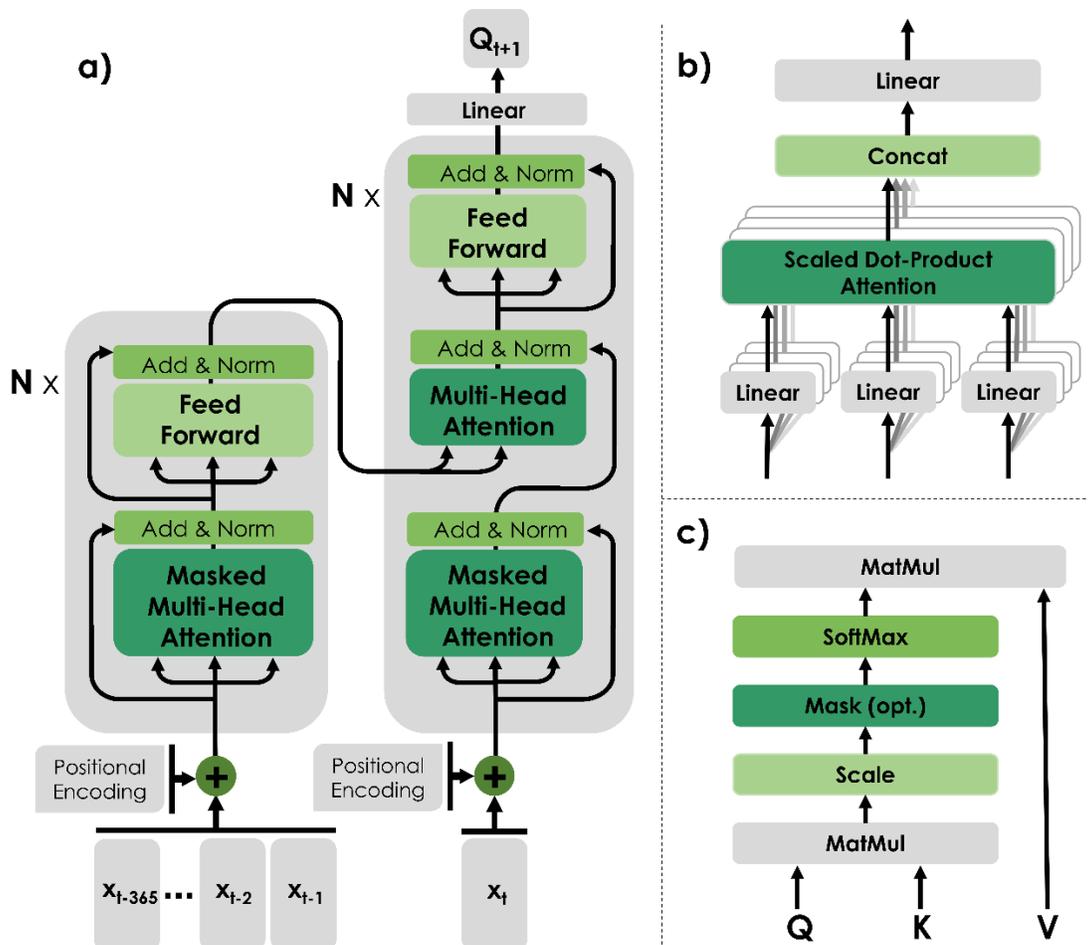

**Figure 3**. The Transformer architecture implemented in our study.

## 2.4. Temporal Fusion Transformers

Although the Transformers can learn from sequential data, they were not developed for time series forecasting as their primary application. This is where the Temporal Fusion Transformer stands out. Borrowing the ideas of multi-head attention from Transformers and the recurrence from RNNs, Lim et al. (2021) developed Temporal Fusion Transformers (TFT) which combines these concepts. A TFT model architecture consists of multiple aligned components, as shown in Figure 4. This includes recurrent networks (LSTM encoder-decoder) and multi-head attention over processed (i.e., after passing through initial components) information of all time steps in the sequence [13]. After processing through variable selection networks (see Section 2.4.2), the ordered sequence is fed into two sets (encoder and decoder) of LSTMs. The past known inputs (such as meteorological forcings) go into the encoder, and known future inputs (such as relative time index in our case) go into the decoder LSTMs. At the current time step $t$, the states of the encoder LSTM are passed into the decoder LSTM. The outputs from these layers are further processed (through gates) and passed into the temporal self-attention layer. Here, the model learns from the entire sequence through multi-head attention, similar to that of Transformers. The output from the attention layer, after being processed through different gates (discussed in later sections), is mapped to seven quantiles of target streamflow through a Dense layer. Additionally, TFT architecture stands out mainly because of three elements, which are (1) gating mechanisms, (2) variable selection networks, and (3) static covariate encoding. These building blocks are discussed in detail in the subsequent sections.

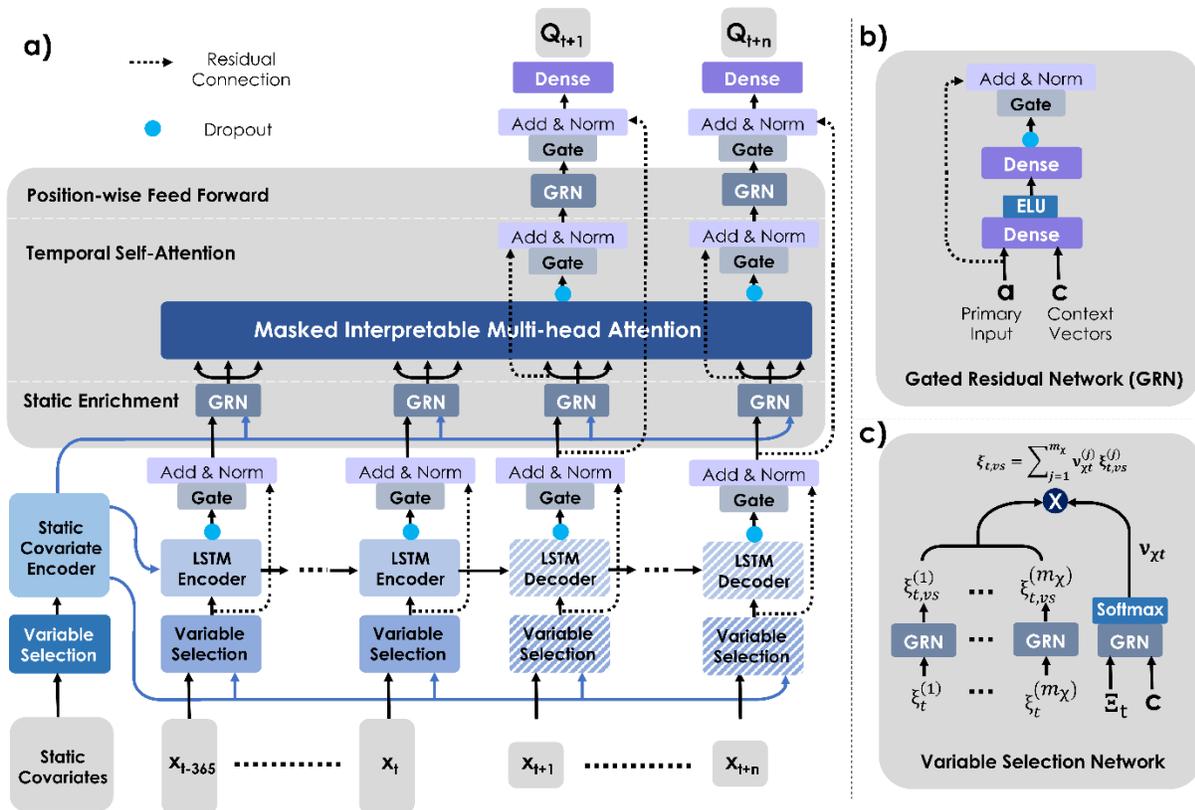

**Figure 4**. The TFT architecture.

### 2.4.1. Gating Mechanisms in TFT

TFT uses Gated Residual Networks (GRNs; Figure 4b) as a key constituent in its architecture. The information flow through the GRNs can be represented using the following equations.

$$GRN(a, c) = LayerNorm(a + GLU(\eta_1)) \tag{9}$$

$$GLU(\eta_1) = \sigma(W_1\eta_1 + b_1) \odot (W_2\eta_1 + b_2) \tag{10}$$

$$\eta_1 = W_3\eta_2 + b_3 \tag{11}$$

$$\eta_2 = ELU(W_4 a + W_5 c + b_4) \tag{12}$$

Here, **a** is the basic input to GRN, and **c** is the optional context vector derived from static covariates. **W** and **b** are weights and biases that are optimized during training. Firstly, **a** and **c** are passed through a dense layer (Equation 12) with Exponential Linear Unit (ELU) activation function [29], followed by a linear layer (Equation 11) with dropout. The output from this layer goes into a Gated Linear Unit [30], represented by Equation 10, where $\odot$ is the element-wise product and **σ** is the sigmoid activation function. Finally, the output from GRN is calculated by standard layer normalization [28] of the sum of input **a** and GLU output (residual connection).

### 2.4.2. Variable Selection Networks in TFT

Regardless of the wide range of input variables available, not all might be relevant for predicting streamflow (or in time series prediction). TFT networks automatically identify important features with significant information about the target variable (streamflow in our case) through its variable selection networks (Figure 4c). Through these networks (or blocks), TFTs can provide interpretable insights into the underlying dynamics of a time series. The following equations can represent the calculations inside a single variable selection block (at time t).

$$v_{\chi t} = \text{Softmax}(GRN(\Xi_t, c_s)) \tag{13}$$

$$\xi_{t,vs}^{(j)} = GRN\left(\xi_t^{(j)}\right) \tag{14}$$

$$\xi_{t,vs} = \sum_{j=1}^{m_\chi} v_{\chi t}^{(j)} \xi_{t,vs}^{(j)} \tag{15}$$

Here, firstly, all variables at time $t$ are linearly transformed to $d_{model}$ dimension. Such transformed input for $j^{th}$ variable is represented by $\xi_t^{(j)}$. Each transformed input is passed through a GRN block (Equation 14), which takes care of the non-linearities. $\xi_{t,vs}^{(j)}$ is the output from these blocks. Along with that, a separate GRN block, which takes the $\Xi_t$ (flattened $\xi_t^{(j)}$) and a context vector $c_s$ derived from static covariates as inputs, produces a vector $v_{\chi t} \in R^{m_\chi}$ after passing through a softmax activation (Equation 13). Note that, $c_s$ is not used in static variable selection block. $v_{\chi t}$ can be interpreted as variable weights assigned for each variable, which are then multiplied with corresponding processed inputs (Equation 15). As shown in Figure 4c, the variable selection is applied for all time steps before further processing. It is important to note that the weights of each GRN block corresponding to all variables and flattened input are shared through all the time steps.

### 2.4.3. Static Covariate Encoding in TFT

Unlike LSTM or Transformers, TFT has dedicated mechanisms to integrate static variables. This is acquired by generating four context vectors from processed and weighted static variables (represented as $\zeta$) generated by the static variable selection network. $\zeta$ is passed through a GRN block to produce four vectors ($c_s$, $c_e$, $c_c$, $c_h$). The generalized form of this calculation can be shown as $c = GRN(\zeta)$. Each context vector is infused into different parts of the TFT, where they are used for processing. For instance, as described in the previous section, $c_s$ is used to find the variable weights inside a variable selection network.

## 2.5. Training

We implement the TFT model in Python using *PyTorch* and *PyTorch Forecasting* libraries. For one basin, we start by organizing the input dataset into a collection of sequences such that one sequence contains all the variables (total 249) of the past 365 days used to predict streamflow of a single day. The dataset is divided into training, validation, and test sets in a 70:10:20 ratio. We train the model using Ranger optimizer [31], a combination of RAdam (Rectified Adam) and LookAhead. This allows fast and efficient optimization. We use the quantile loss function, as TFT can produce quantile outputs through linear mapping to the desired number of quantiles. Along with a batch size of 32 and a maximum of 100 epochs, we use early stopping to avoid overfitting and faster convergence. These training configurations are used based on several iterations of the model on a number of randomly selected basins. In the case of Transformers, we use the same training configurations. The only difference is that the data for the first 364 days is fed into the encoder, and the data for the 365th day goes into the decoder part of Transformers. For LSTMs, we use the mean-squared error loss function for training, with the remaining configurations the same as that of TFT. For all the 2,610 basins, we separately train models with the same configurations for each model, as discussed above. The computational cost for the TFT model is relatively high since we have a large number of input variables. A TFT model for one basin takes one hour to train with 2 GPU cores. Finally, we evaluate the performance of all the models with the Kling-Gupta Efficiency (KGE) [32].

## 3. Results and Discussion

In this section, we compare the performance scores produced by the LSTM, Transformers, and TFTs. This provides evidence in support of our hypothesis that combining recurrence with attention improves streamflow prediction. The discussion continues by juxtapositioning three hydrographs (generated by each model) of one basin in North America, showing the information about streamflow learned by the model from the data. Further, we discuss how well TFT is able to learn the long-term dependencies in the data.

### 3.1. Performance Comparison

Figure 5 shows the histogram of KGE scores by three models from the test data of 2,610 basins globally. The median KGEs are 0.647, 0.563, and 0.705 for LSTMs, Transformers, and TFTs, respectively. Clearly, the TFT, a combination of recurrent and attention-based networks, outperforms the popular LSTM and Transformers in streamflow prediction. This validates our hypothesis. Moreover, the frequency of basins with near 1.0 KGE scores is much higher in TFT than in the other two models. With the help of the hydrographs of a single basin (hysets_10BE007) generated by three models, as shown in Figure 6, we discuss the possible reasons for the better performance of TFTs. Comparing these hydrographs, TFT is superior in terms of capturing the peak discharge, rising limbs, and falling limbs. The potential reason why TFT outperforms LSTM is that TFT, for predicting the peak and limbs at a time, can concurrently learn from all other peaks and limbs in the hydrograph. The LSTM, since it sequentially learns from previous time steps, might not be capturing information from the distant past regardless of its memory. Although this ability exists in a Transformer, it cannot perform as an LSTM, most likely because Transformer architecture is not representative of dynamical processes in catchments. That is, the hydrologic processes in a catchment occur in a sequential manner, step-by-step, similar to the data processing inside an LSTM. Additionally, the local context, such as soil moisture changes or the cyclic nature of meteorological forcings (e.g., temperature), is important for streamflow predictions. The capabilities of a Transformer to learn these local contexts might be limited, as it tries to learn from every point in a sequence individually (point by point). By combining aspects of LSTM and Transformers, TFT can efficiently extract knowledge from near and distant data, thereby outperforming the former two models. However, further improving the performance through rigorous hyperparameter tuning is possible. In this study, we have not tuned the hyperparameters for each basin separately, as it is computationally extremely expensive.

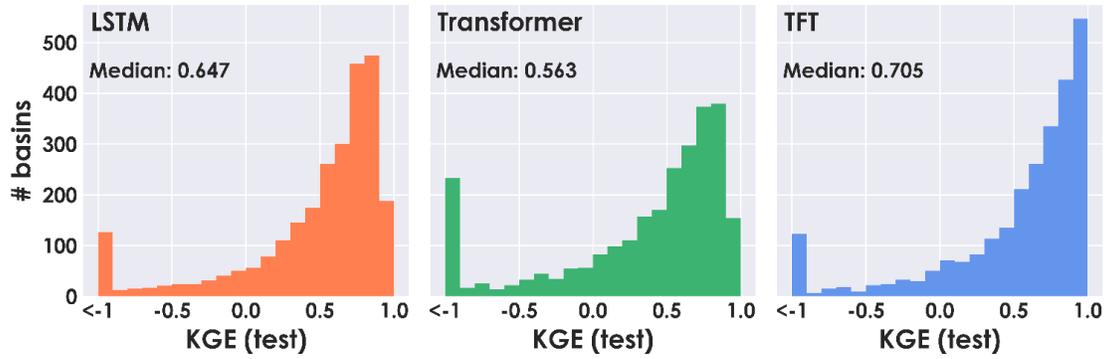

**Figure 5.** Distribution of test case KGE scores of observed and model-predicted streamflow. The values below the negative one are put into the first bin for better visualization.

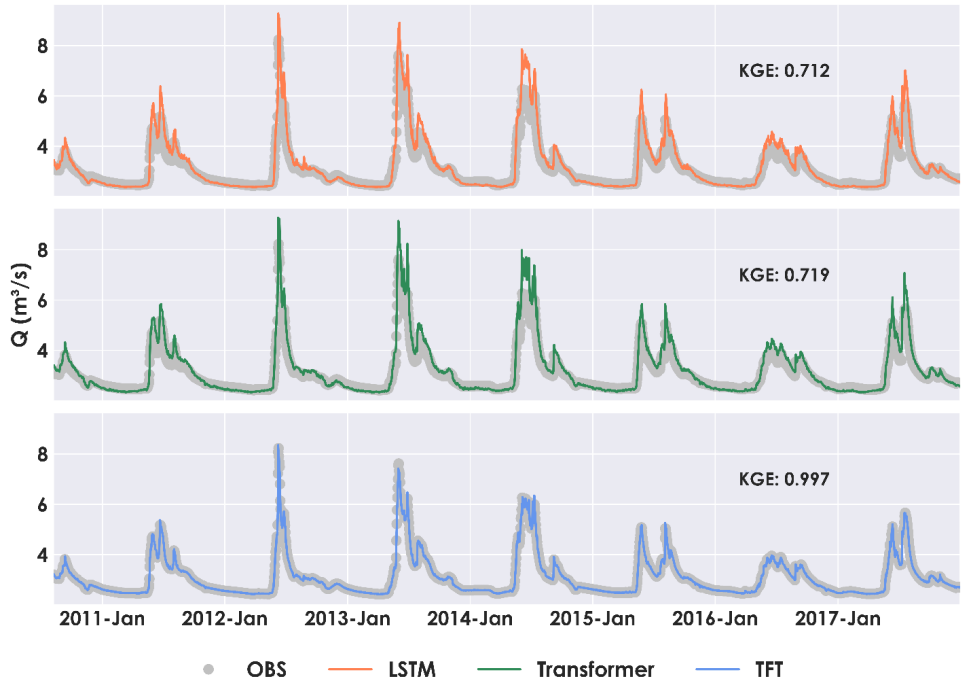

**Figure 6.** Hydrograph generated by three models for station hysets_10BE007.

## 3.2. Model Interpretability

As discussed earlier, one of the key features for which TFT excels over other networks is the interpretability of the model. TFT achieves this through its variable selection networks. To demonstrate how a TFT model can give insights into the hydrological processes, we plotted (Figure 7) the relative feature importance of the basin hysets_10BE007 (same basin discussed in the previous subsection) provided by the variable selection networks of TFT. Among continuous variables, past streamflow, total precipitation, water contents in soil layers, and potential evaporation are weighted as the top four variables by TFT for this basin. In static variables, monthly mean potential evapotranspiration (PET) in November has the most importance, followed by the minimum net surface solar radiation, frequency of precipitation days, and land cover extents. This outcome is quite intuitive and complies well with the domain knowledge. Streamflow being the target variable, past values contain significant information (considerably more than any other variable) about the patterns in it. Total precipitation, the highest weighted exogenous variable, directly drives the streamflow generation in the basins. The same reason can be used to explain why the model gives importance to the frequency of precipitation days, a static covariate. The water content in soil layers can drive surface runoff and infiltration directly. In the case of this particular basin, total potential evaporation and monthly mean PET in November (static) have a significant impact on streamflow. In light of these results, a basin-specific investigation can help us understand why potential evaporation, especially in November, is important in determining streamflow in the basin. However, this might not be the case in all other basins. For instance, in basins where snow processes have a vital role in runoff generation, snow water equivalent might surpass other variables in terms of importance. Further analysis of region-wise interpretation of TFTs is required to study such effects, and it can potentially provide new knowledge about the hydrological processes of different regions.

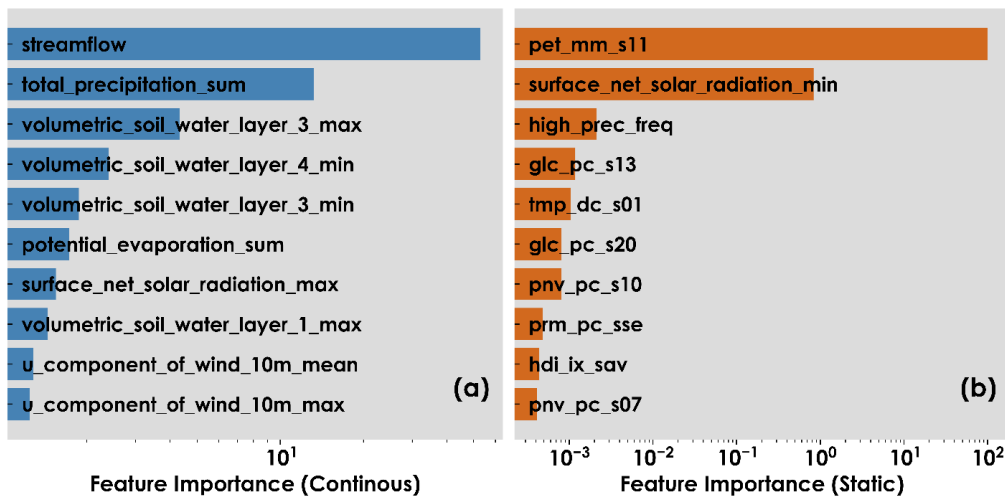

**Figure 7.** Variable importance identified by TFT for basin hysets_10BE007. For more information about the variables, please refer to Kratzert et al. (2023).

### 3.3. Learning Long-term Patterns

Arguably, the most important reason TFTs outperform LSTM is their ability to access and learn from data of the distant past directly. For instance, the streamflow pattern in a given week of any year can be closely related to the streamflow during the same week in the previous year due to the effect of seasonality in the hydrologic processes. Therefore, a model can possibly learn vital information about streamflow from the previous year's data. We demonstrate the ability of TFT to learn seasonal dependencies through attention scores of trained models over a sequence of the past year. Figure 8 shows the distribution of percentage attention (over the sequence from *t-365* till *t*) given by the trained TFTs of all basins. The values are plotted by compartmentalizing them into months, resulting in 12 bins of past months (lookback). It is evident from the plot that the TFTs give higher attention scores for time steps of the immediate past (1-month lookback), most likely learning the near-term context. Importantly, TFTs also give high attention scores for the time steps in the distant past (12-month lookback), most likely learning from the seasonal dependencies of hydrologic processes embedded in the data. In this study, we trained all models with sequences of the past 365 days. However, the performance of attention-based models, especially TFT, can grasp more knowledge from sequences of multi-year length because the model can attend to and learn from multiple seasons in the sequence and longer-term temporal patterns (e.g., decadal variability).

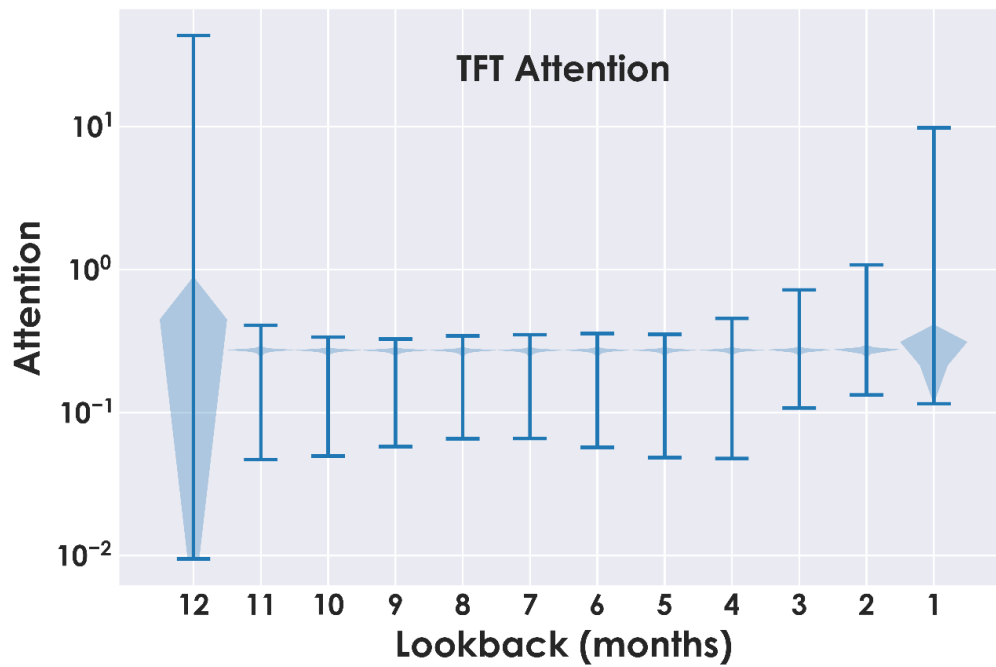

**Figure 8.** Distribution of relative attention scores by TFT for previous time steps.

## 4. Conclusion

This study tests the hypothesis that combining attention with recurrence improves streamflow prediction. We train three state-of-the-art machine learning models - LSTM, Transformer, and TFT – for 2,610 basins worldwide separately to predict the streamflow. The basins included in this study are selected from the Caravan dataset, filtering those without significant missing data. We compare the performance scores and hydrographs generated by the three models.

Our results show that TFT, a model combining the aspects of recurrence from LSTM and attention from Transformers, outperforms the latter two. The TFT is able to capture the nuances in the hydrographs, including the peaks and limbs, more efficiently than LSTM or Transformers. We demonstrate that the TFT can learn long-term dependencies through its attention mechanism. To explain the interpretability of TFT, we discuss the variable importance identified by it that can shed light on unknown information about hydrological processes in basins. This study introduces the implementation of TFT in hydrology and demonstrates how that outperforms the current state-of-the-art ML models and sets a new benchmark.

## Conflicts of Interest

The authors express no conflict of interest.